\documentclass[runningheads]{llncs}
\usepackage{graphicx}
\usepackage{amsmath}
\usepackage{algorithm}
\usepackage{algorithmic}
\usepackage{bbm}
\usepackage{hyperref}
\usepackage{url}
\usepackage{booktabs}
\usepackage{amsfonts}
\usepackage{nicefrac}
\usepackage{microtype}
\usepackage{xcolor}
\usepackage{lipsum}
\usepackage{multirow}
\usepackage{caption}
\usepackage{subcaption}
\usepackage[misc]{ifsym}
\begin{document}
\title{Adaptation of Surgical Activity Recognition Models Across Operating Rooms}
\titlerunning{Adaptation of Surgical Activity Recognition}
% If the paper title is too long for the running head, you can set
% an abbreviated paper title here
%
\author{Ali Mottaghi \inst{1, 2} \and
Aidean Sharghi \inst{1} \and
Serena Yeung \inst{2} \and
Omid Mohareri \inst{1}}
% index{Mottaghi, Ali}
% index{Sharghi, Aidean}
% index{Yeung, Serena}
% index{Mohareri, Omid}
%
\authorrunning{A. Mottaghi et al.}
% First names are abbreviated in the running head.
% If there are more than two authors, 'et al.' is used.
%
\institute{Intuitive Surgical Inc., Sunnyvale, CA \and 
Stanford University, Stanford, CA \\
\email{mottaghi@stanford.edu}}
% \institute{Paper ID: 2509}
%
\maketitle              % typeset the header of the contribution
\begin{abstract}
Automatic surgical activity recognition enables more intelligent surgical devices and a more efficient workflow. Integration of such technology in new operating rooms has the potential to improve care delivery to patients and decrease costs. Recent works have achieved a promising performance on surgical activity recognition; however, the lack of generalizability of these models is one of the critical barriers to the wide-scale adoption of this technology. In this work, we study the generalizability of surgical activity recognition models across operating rooms. We propose a new domain adaptation method to improve the performance of the surgical activity recognition model in a new operating room for which we only have unlabeled videos. Our approach generates pseudo labels for unlabeled video clips that it is confident about and trains the model on the augmented version of the clips. We extend our method to a semi-supervised domain adaptation setting where a small portion of the target domain is also labeled. In our experiments, our proposed method consistently outperforms the baselines on a dataset of more than 480 long surgical videos collected from two operating rooms.

\end{abstract}

\section{Introduction}
Surgical workflow analysis is the task of understanding and describing a surgical process based on videos captured during the procedure. Automatic surgical activity recognition in operating rooms provides information to enhance the efficiency of surgeons and OR staff, assess the surgical team's skills, and anticipate failures \cite{vercauteren2019cai4cai}. Video activity recognition models have been previously utilized for this task; however, one of the main drawbacks of previous works is the lack of generalizability of the trained machine learning models. Models trained on videos from one operating room perform poorly in a new operating room with a distinct environment \cite{sharghi2020automatic} \cite{schmidt2021multi}. This prevents the widespread adoption of these approaches for scalable surgical workflow analysis. 

This paper explores approaches for adapting a surgical activity recognition model from one operating room to a new one. We consider two cases where we have access to only unlabeled videos on the target operating room and the case where a small portion of target operating room videos is annotated. Our new approach generates pseudo labels for unlabeled video clips on which the model is confident about the predictions. It then utilizes an augmented version of pseudo-annotated video clips and originally annotated video clips for training a more generalizable activity recognition model. Unlike most previous works in the literature that study semi-supervised learning and domain adaptation separately, we propose a unified solution for our semi-supervised domain adaptation problem. This allows our method to exploit labeled and unlabeled data on the source and the target and achieve more generalizability on the target domain. 

Our work is the first to study semi-supervised domain adaptation in untrimmed video action recognition. Our setting of untrimmed video action recognition (1) requires handling video inputs, which limits the possible batch size and ability to estimate prediction score distributions as needed by prior image classification works \cite{berthelot2021adamatch}; and (2) requires handling highly imbalanced data, which causes model collapse when using the pseudo labeling strategies in prior image classification works. Our method addresses the first challenge by introducing the use of prediction queues to maintain better estimates of score distributions and the second challenge through pretraining and sampling strategies of both video clips and pseudo labels to prevent model collapse. Additionally, we use a two-step training approach as described in section~\ref{sec:untrimmed}.
% by first training a clip-based backbone model on the videos and then training a temporal model on features extracted from the backbone action recognition model. 

A dataset of more than 480 full-length surgical videos captured from two operating rooms is used to evaluate our method versus the baselines. As we show in our experiments, our new method outperforms existing domain adaptation approaches on the task of surgical action recognition across operating rooms since it better utilizes both annotated and unannotated videos from the new operating room.

\section{Related Works}
\textbf{Surgical Activity Recognition} also known as surgical phase recognition, is the task of finding the start and end time of each surgical action given a video of a surgical case. \cite{tran2017phase} \cite{yengera2018less} \cite{zia2018surgical} This task has been studied for both laparoscopic videos \cite{yengera2018less} \cite{zia2018surgical} \cite{funke2018temporal} \cite{chen2018endo3d} and videos captured in the operating room \cite{sharghi2020automatic} \cite{schmidt2021multi}. These videos are often hours long, containing several activities with highly variable lengths. Unlike trimmed video action recognition, which is studied extensively in computer vision literature, untrimmed video action recognition does not assume that the action of interest takes nearly the entire video duration. Despite the relative success of surgical activity recognition models, previous works \cite{sharghi2020automatic} \cite{schmidt2021multi} have shown that these models suffer from the domain shift as a model trained in one OR performs significantly worse in a new OR. In this work, we address this problem by proposing a new method for training a more generalizable model with minimal additional annotations from the new OR. 

\textbf{Unsupervised Domain Adaptation (UDA)} is the problem of generalizing the model trained on the source domain to the target domain where we only have unlabeled data. The goal is to achieve the highest performance on the target domain without any labeled examples on target. Most of the works in UDA literature focus on reducing the discrepancy of representations between source and target domains. For example, \cite{long2015learning} and \cite{long2016unsupervised} use the maximum mean discrepancy to align the final representations while \cite{ganin2016domain} \cite{sun2016deep} suggest matching the distribution of intermediate features in deep networks. Maximum classifier discrepancy (MCD) \cite{saito2018maximum} has been proven more successful where two task classifiers are trained to maximize the discrepancy on the target sample. Then the features generator is trained to minimize this discrepancy. 

\textbf{Semi-supervised Domain Adaptation (SSDA)} setting has access to a labeled set on the target domain, unlike UDA. The goal is still achieving the highest performance on the target domain, however, with the use of both labeled and unlabeled data on source and target. SSDA is less explored than DA, and most of the works have focused on image classification tasks \cite{yao2015semi} \cite{ao2017fast} \cite{saito2019semi}. For example, in Minimax Entropy (MME) \cite{saito2019semi} approach, adaptation is pursued by alternately maximizing the conditional entropy of unlabeled target data with respect to the classifier and minimizing it with respect to the feature encoder. Recently, \cite{berthelot2021adamatch} suggested AdaMatch, where the authors extend FixMatch to the SSDA setting. Our work borrows some ideas from this work, but unlike AdaMatch, we focus on video action recognition. We also proposed a new method for addressing the long-tailed distribution of data on the target domain, which is very common in surgical activity recognition. 

\section{Method}

\subsection{Unsupervised Domain Adaptation (UDA)}
\label{sec:uda}
\textbf{Notation.} Let $\mathcal{V}_{s} = \{v_{s}^{(1)},\dots, v_{s}^{(n_{s})}\}$ be the set of videos in the source domain and $\mathcal{V}_{t} = \{v_{t}^{(1)}, \dots, v_{t}^{(n_{t})}\}$ be videos on target where $n_{s}$ and $n_{t}$ are the number of source and target videos. As is common in UDA, we assume source data is labeled while the data from the target is unlabeled. Denote the set of surgical activities as $\mathcal{C} = \{c_1, \dots, c_K\}$ where $K$ is the number of surgical activities of interest. For each source video $v_{s}^{(i)}$, we are given a set of timestamps $s_{s}^{(i)}=\{t_{c_1}, \dots, t_{c_k}\}$ where $t_{c_j}$ is defined as the start and end time for activity $c_j$. 

\textbf{Sampling Video Clips.} We start with sampling short video clips from long and variable length videos. Let $x_{s} \in \mathbb{R}^{H \times W \times 3 \times M}$ denote a sample clip from source videos and $x_{t}$ be a sample clip from the target. All sampled clips have a fixed height $H$, width $W$, and number of frames $M$. We sample clips uniformly from surgical activities for labeled videos to ensure each clip only contains one activity. Let $y_{s} \in \mathcal{C}$ be the corresponding label for $x_{s}$. We don't have any labels for target clip $x_{TU}$.

\textbf{Augmentations.} In our method, we feed both the sampled video clip $x$ and an augmented version $\tilde{x}$ to our model. Augmentation is performed both frame-by-frame and temporally. For frame-wise augmentation, we use RandAugment \cite{cubuk2020randaugment}, and we change the playback speed for temporal augmentation. More formally, with probability $p$, the video clip is up-sampled or down-sampled by a factor $\alpha$. The activity recognition model $F$ takes an input video clip $x$ and outputs logits $z \in \mathbb{R}^{K}$ for each surgical activity. For augmented clip from source we have $\tilde{z}_{s} = F(\tilde{x}_{s})$ and $\tilde{p}_{s} = softmax(\tilde{z}_{s})$ where $\tilde{p}_{s}$ is a vector of predicted probabilities. Similarly, we define $z_{s}$, $z_{t}$, $\tilde{z}_{t}$, $p_{s}$, $p_{t}$, and $\tilde{p}_{t}$.

\begin{figure}[t]
\begin{center}
    \includegraphics[width=0.9\textwidth]{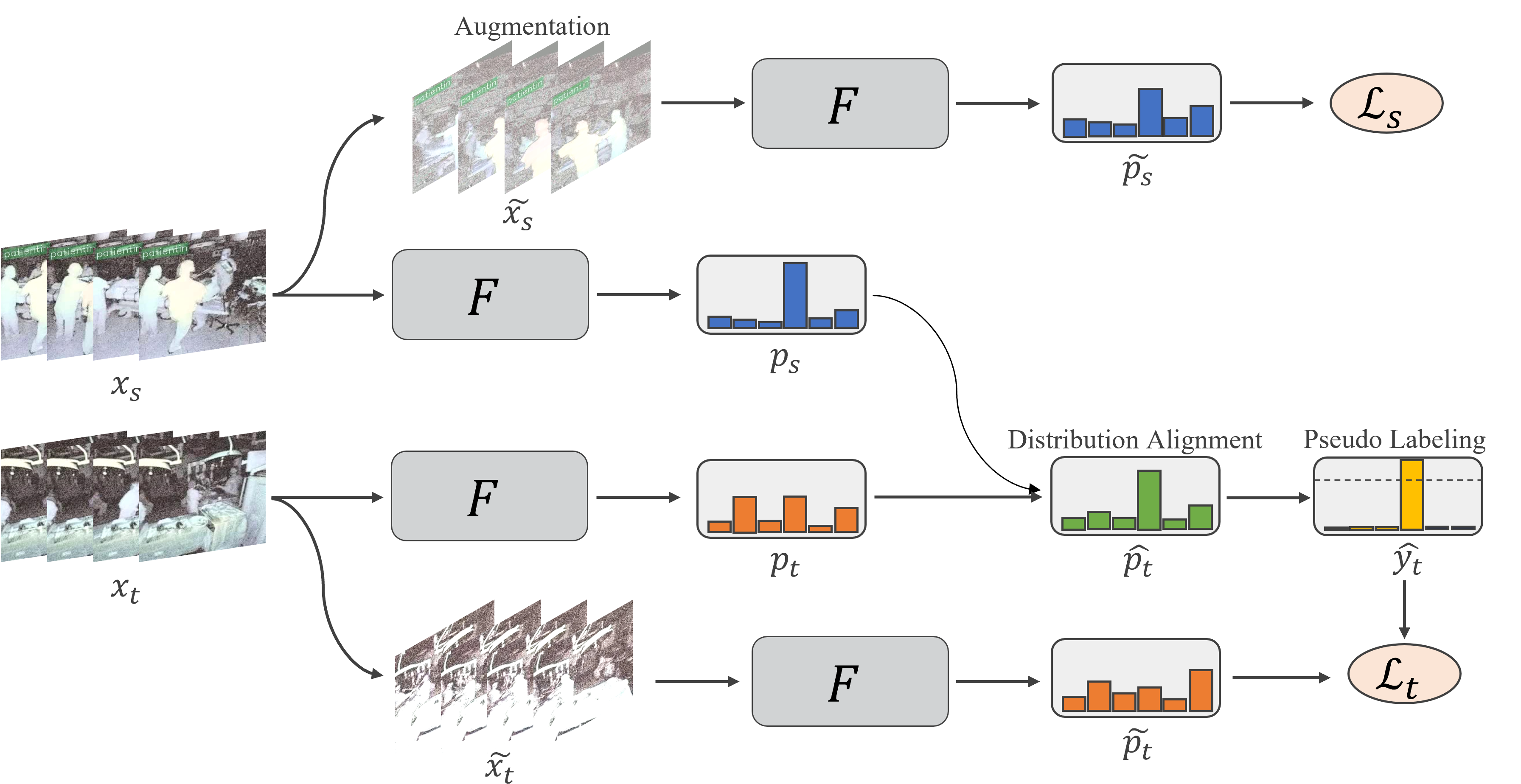}
\end{center}
\caption{Our proposed domain adaptation method. First, the labeled video clip from the source $x_s$ and the unlabeled video clip form the target $x_t$ are augmented to create $\tilde{x}_{s}$ and $\tilde{x}_{t}$ respectively. Then all four video clips are fed into the clip-based activity recognition model $F$ to generate probability distributions $p_s$, $\tilde{p}_{s}$, $p_t$, and $\tilde{p}_{t}$. For the source $x_s$, we already have the label, so we compute the $\mathcal{L}_s$. For the target, we generate and sample a pseudo label as described in section \ref{sec:uda} before computing $\mathcal{L}_t$. Note that the loss is computed only for the augmented version of the video clips to improve generalizability.
}
\label{fig:method}
\end{figure}

\textbf{Distribution Alignment.} Before generating pseudo labels for unlabeled clips from target, we align the distributions of predictions on source and target domains by aligning their expected values. Unlike AdaMatch \cite{berthelot2021adamatch} that uses only one batch to estimate the source and target distributions, we estimate them by maintaining queues of previous predictions $\mathcal{P}_{s} = \{p_{s}^{(0)}, p_{s}^{(1)}, \dots \}$,  $\mathcal{P}_{t} = \{p_{t}^{(0)}, p_{t}^{(1)}, \dots \}$. $\mathcal{P}_{s}$ and $\mathcal{P}_{t}$ are defined on the fly as queues with the current mini-batch enqueued and the oldest mini-batch dequeued. The size of the queues can be much larger than mini-batch size and can be independently and flexibly set as a hyperparameter. This is particularly useful in our setting where the typical mini-batch size is small due to GPU memory restrictions. We calculate the aligned prediction probabilities for the target by $\hat{p}_{t} = normalize(p_{t} \times \bar{p}_{s}/\bar{p}_{t})$ where $\bar{p}_{s} = \mathbb{E}_{p_{s} \sim \mathcal{P}_{s}} [p_{s}]$ and $\bar{p}_{t} = \mathbb{E}_{p_{t} \sim \mathcal{P}_{t}} [p_{t}]$. This ensures that despite the difference in input distribution of source and target, the output predictions of the model are aligned.

\textbf{Pseudo Labeling.} We generate pseudo labels for the most confident predictions of the model. In this way, the model can be trained on the target video clips that are originally unlabeled. If the maximum confidence in the target prediction is less than $\tau$, the target sampled clip will be discarded, i.e. $mask = \max(\hat{p}_{t}) > \tau$, otherwise, a pseudo label is generated as $\hat{y}_{t} = \arg \max(\hat{p}_{t})$.

\textbf{Sampling Pseudo Labels.} In practice, the generated pseudo labels have an imbalanced distribution as the model is usually more confident about the dominant or easier classes. If we train the model on all generated pseudo labels, the classifier could predict the most prevalent class too often or exhibit other failure modes. We only use a subset of pseudo labels based on their distribution to mitigate this problem. We maintain another queue $\mathcal{Y}_{t} = \{\hat{y}_{t}^{(0)}, \hat{y}_{t}^{(1)}, \dots \}$ to estimate the distribution of generated pseudo labels. Let $Q$ be the vector of frequencies of pseudo labels in $\mathcal{Y}_{t}$ where $Q_i$ is the number of repeats for class $i$. We sample generated pseudo labels with the probability proportional to the inverse of frequency of that class, i.e. $sample \sim bernoulli(\min(Q) / Q_{\hat{y}_{t}})$. This adds more representation to the hard or infrequent classes on target. 

\textbf{Loss Functions.} The overview of our domain adaptation method is depicted in Fig. \ref{fig:method}. To sum up the discussion on UDA, we define loss function $\mathcal{L}$:
\begin{align*}
    \mathcal{L}_{s} &= \mathbb{E}_{x_s \sim X_s, y_s \sim Y_s} [H(y_s, \tilde{z}_{s})] \\
    \mathcal{L}_{t} &= \mathbb{E}_{x_s \sim X_s, x_t \sim X_t, y_s \sim Y_s} [H(\text{stopgrad}(\hat{y}_{t}), \tilde{z}_{t}) \cdot mask \cdot sample]  \\
    \mathcal{L} &= \mathcal{L}_{s} + \lambda \mathcal{L}_{t}
\end{align*}
Where $H$ is the categorical cross-entropy loss function and $\text{stopgrad}$ stops the gradients from back-propagating. $\lambda$ is a hyperparameter that weighs the domain adaptation loss. We use a similar scheduler as \cite{berthelot2021adamatch} to increase this weight during the training. Note that the augmented video clips are utilized during the back-propagation, and the original video clips are used only to generate pseudo labels.

\begin{figure}[t]
\begin{center}
    \includegraphics[width=0.7\textwidth]{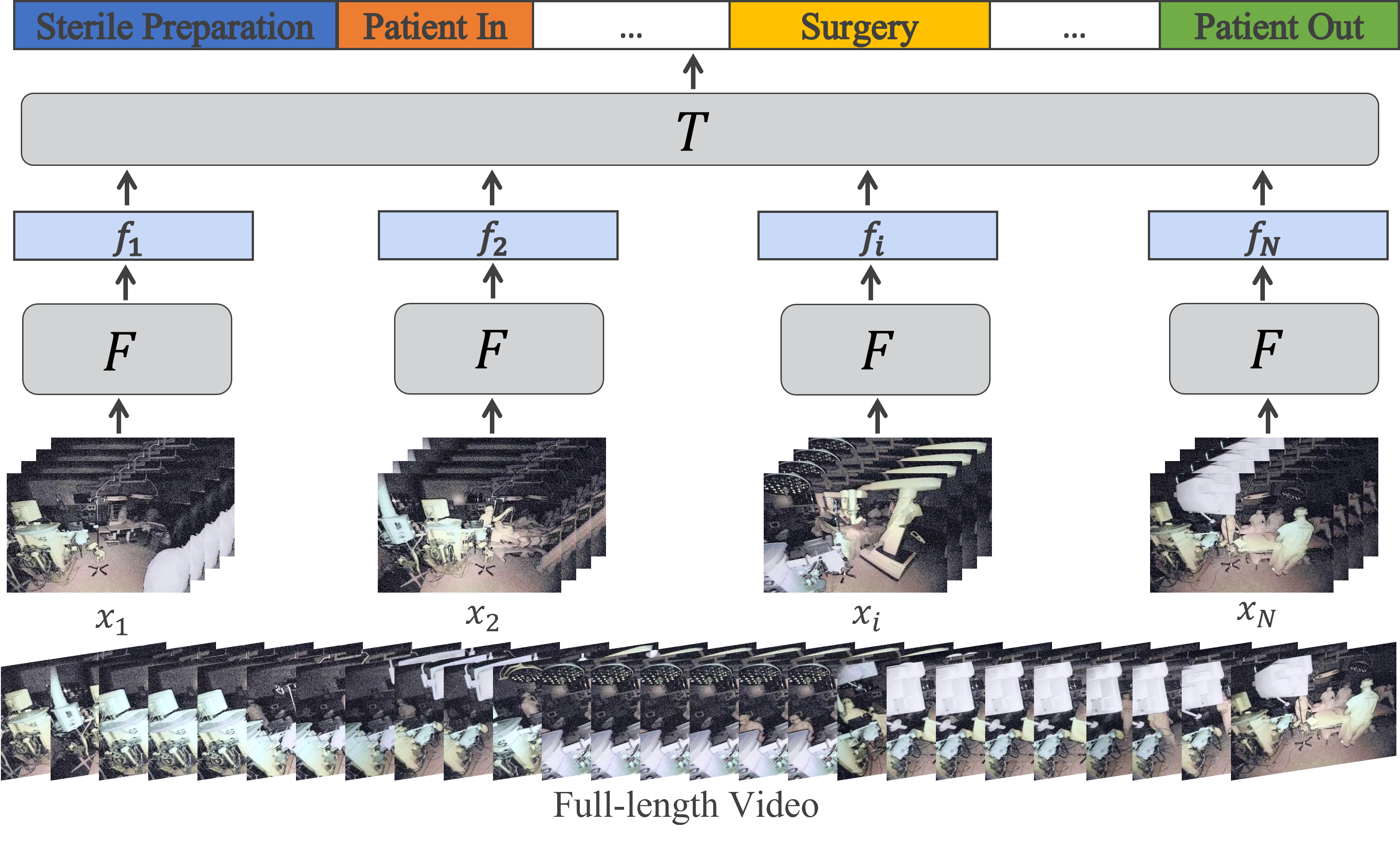}
\end{center}
\caption{The original operating room videos are usually long, so we split them into short video clips ($x_1, x_2, \dots, x_N$) and feed them into a trained clip-based activity recognition model $F$. Next, the extracted features ($f_1, f_2, \dots, f_N$) are used to train a temporal model $T$, which predicts the surgical activity of each frame in the original video given the temporal context.}
\label{fig:temporal}
\end{figure}

\subsection{Training on Untrimmed Surgical Videos}
\label{sec:untrimmed}
Clip-based activity recognition models cannot model long-term dependencies in long untrimmed videos. However, surgical videos captured in operating rooms usually have a natural flow of activities during a full-length surgery. We train a temporal model on top of the clip-based activity recognition model to capture these long-term dependencies. As depicted in Fig. \ref{fig:temporal}, first, we extract features from all videos by splitting long videos into clips with $M$ frames and feeding them into the trained clip-based model. Features are extracted from the last hidden layer of model $F$. Let $f \in \mathbb{R}^{D \times N}$ be the features extracted from video $v \in \mathbb{R}^{H \times W \times 3 \times N}$ and $l \in \mathbb{R}^{K \times N}$ be the corresponding ground-truth label, where $N$ is the number of frames in the video and $D$ is the latent dimension of the clip-based model. We train the temporal model by minimizing $\mathcal{L}_{temporal} = H(l, \hat{l})$ where $\hat{l} = T(f)$ is the prediction and $H$ is the binary cross-entropy loss. 

\subsection{Extension to Semi-supervised Domain Adaptation (SSDA)}
Our domain adaption method can be extended to SSDA setting. In SSDA, we assume the source videos $\mathcal{V}_{s}$ are fully labeled, while only a part of target videos $\mathcal{V}_{tl}$ are labeled; the rest are denoted by $\mathcal{V}_{tu}$ are unlabeled. Therefore, in addition to sampling video clips $x_{s}$ and $x_{tu}$ as described in section 3.1, we also sample video clips $x_{tl}$ from target labeled videos. This adds an additional term $\mathcal{L}_{tu}$ to the loss function: 
\begin{align*}
    \mathcal{L}_{tl} &= \mathbb{E}_{x_{tl} \sim X_{tl}, y_{tl} \sim Y_{tl}} [H(y_{tl}, \tilde{z}_{tl})]
\end{align*}
The total loss function is $\mathcal{L} = \mathcal{L}_{s} + \mathcal{L}_{tl} + \lambda \mathcal{L}_{tu}$ where $\mathcal{L}_{s}$ and $\mathcal{L}_{tu}$ are defined as before. Our distribution alignment and pseudo labeling strategies are similar to UDA setting. 

\subsection{The Importance of Pretraining}
\label{sec:pre}
We pretrain our activity recognition model to boost its performance in domain adaptation training. More specifically, in the beginning, we train the activity recognition model only on source data by optimizing $\mathcal{L}_s$. This has several advantages: 1) Predicted probabilities are more meaningful since the beginning of the training. As a result, generated pseudo labels will be more accurate and reliable. 2) Distributions are estimated more accurately; therefore, we can better align distributions and sample more representative pseudo labels. 3) If we train the temporal model on source data, we can also use its predictions on unlabeled targets for sampling video clips. Videos are usually long, containing multiple activities with highly variable lengths. Therefore, we sample a more uniform set of video clips covering all surgical activities with an approximate segmentation of surgical activities. We achieve this by first sampling uniformly from surgical activities and then sampling short video clips from each.

\section{Experiments}

\textbf{Dataset.} We use a dataset of 484 full-length surgery videos captured from two robotic ORs equipped with Time-of-Flight sensors. The dataset covers 28 types of procedures performed by 16 surgeons/teams using daVinci Xi surgical system. 274 videos are captured in the first OR, and the remaining 210 videos are from the second OR with a distinct layout and type of procedures and teams. These videos are, on average, 2 hours long and are individually annotated with ten clinically significant activity classes such as sterile preparation, patient roll-in, etc. Our classes are highly imbalanced; patient preparation class contains ten times more frames than robot docking class.  

\textbf{Model Architectures and Hyperparameters.} We use a TimeSformer \cite{bertasius2021space} model with proposed hyperparameter as our clip-based activity recognition model which operates on $224 \times 224 \times 3 \times 16$ video clips. Clips are augmented with RandAugment \cite{cubuk2020randaugment} with the magnitude of $9$ and standard deviation of $0.5$, as well as temporal augmentation with a factor of $2$ and probability of $0.5$. Given the size of our dataset, we use a GRU \cite{chung2014empirical} as our temporal model. We generate pseudo labels with $\tau = 0.9$ and keep a queue of $1000$ previous predictions and pseudo labels for distribution alignment and sampling. We set $\lambda=1$ in our experiments. The sensitivity analysis of hyperparameters and further implementation details of our approach are available in the supplementary material. 

\textbf{Baselines.} We compare the performance of our domain adaptation strategy to three baselines: 1)\textit{Supervised} method serves as a fundamental baseline where only labeled data is used during the training. 2)\textit{Maximum Classifier Discrepancy (MCD)} \cite{saito2018maximum} is a well-known domain adaption method that has been applied to various computer vision tasks. 3)\textit{Minimax Entropy (MME)} \cite{saito2019semi} is specially tailored to the SSDA settings where adversarial training is used to align features and estimate prototypes. Our method borrows some components from AdaMatch \cite{berthelot2021adamatch}, and we study the effect of each component in our ablation studies. 

\begin{table}[t]
\caption{UDA experiments. For each method, we report the accuracy of the clip-based model and mAP of the temporal model.
}
\centering
\begin{tabular}{l | c c c c}
\toprule[1pt]
\multirow{2}{*}{Method} & \multicolumn{2}{c}{OR1 to OR2} & \multicolumn{2}{c}{OR2 to OR1} \\
 & Accuracy & mAP & Accuracy & mAP  \\ \hline
Source Only & 62.13 & 76.39 & 66.99 & 86.72 \\
MCD \cite{saito2018maximum} & 63.29 & 76.89 & 65.17 & 86.07 \\
MME \cite{saito2019semi} & 67.87 & 81.20 & 68.02 & 88.96 \\
Ours & \textbf{70.76} & \textbf{83.71} & \textbf{73.53} & \textbf{89.96} \\
\bottomrule[1pt]
\end{tabular}
\label{table:UDA}
\end{table}

\textbf{Evaluation.} The performance of each method is evaluated based on the accuracy of the clip-based activity recognition model and the mean average precision (mAP) of the temporal model on the target domain. The accuracy of the activity recognition model evaluated on class-balanced sampled video clips acts as a proxy for the informativeness of the extracted features and is more aligned with most prior works focused on training a classifier. In all experiments, we start from a pre-trained model as discussed in \ref{sec:pre}.

\textbf{UDA Experiments.} Table \ref{table:UDA} shows the performance of our method compared to the baselines in UDA setting. We measure the performance of both the clip-based model (with accuracy) and the temporal model (with mAP). As experiments show, our method outperforms all baselines in both scenarios:  when OR1 is source, and we want to adapt to OR2 as our target and vice versa. The performance is higher on the task of adapting OR2 to OR1 since OR2 includes unique surgical procedures that are not conducted in OR1. For the rest of the section, we only consider adaptation from OR1 to OR2. 

\begin{figure}[tbp]
\begin{subfigure}{0.49\hsize}
    \begin{center}
        \includegraphics[width=1\hsize]{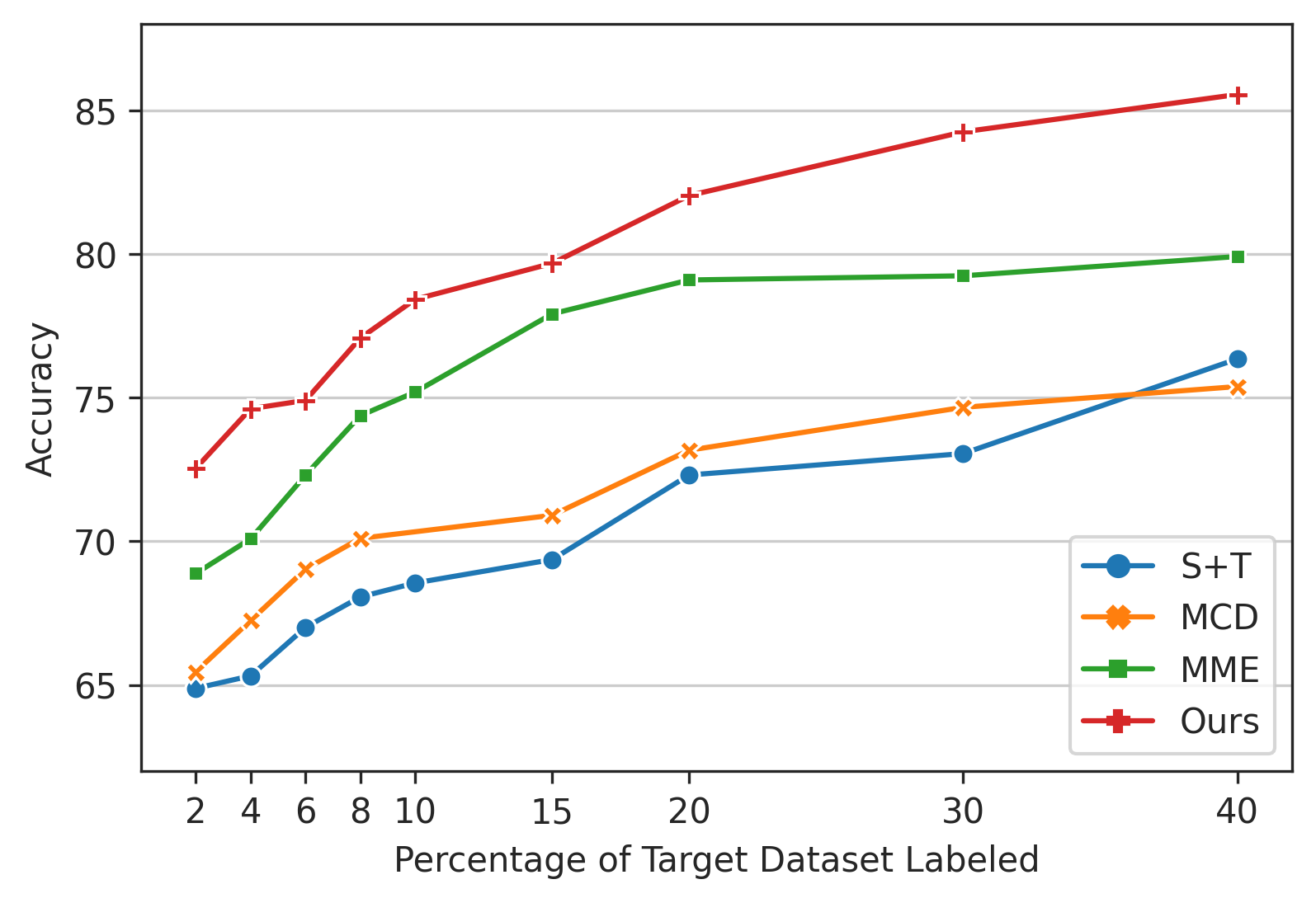}
        % \caption{Clip-based Activity Recognition Model}
    \end{center}
\end{subfigure}
\begin{subfigure}{0.49\hsize}
    \begin{center}
        \includegraphics[width=1\hsize]{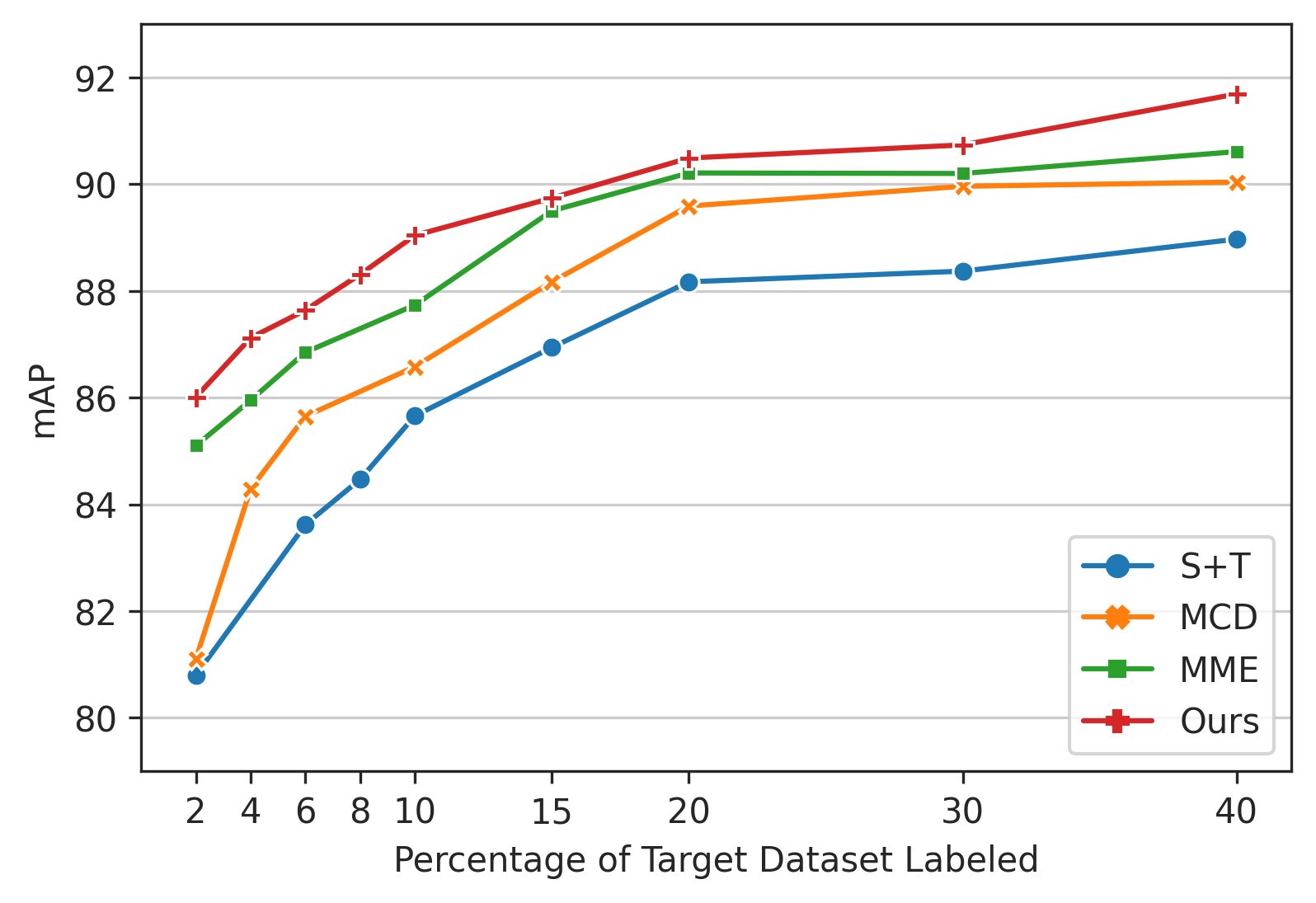}
        % \caption{Temporal Model}
    \end{center}
\end{subfigure}
\caption{Performance of different methods in the SSDA setting. The left figure shows the accuracy of the clip-based activity recognition model, and the right figure shows the corresponding mAP of the temporal model trained on extracted features. S+T denotes training a supervised model on the source plus the labeled part of the target dataset.}
\label{fig:SSDA}
\end{figure}

\textbf{SSDA Experiments.} In SSDA setting, a random subset of the target dataset is chosen to be labeled. Figure \ref{fig:SSDA} shows the performance of our method compared to baselines as we vary the ratio of the target dataset that is labeled. Our method outperforms all of the baselines on our surgical activity recognition task. 

\begin{table}[t]
\caption{Ablation study on two key components of our model. 
% Both distribution alignment and sampling pseudo labels help the model generate better pseudo labels for training.
}
\centering
\begin{tabular}{l | c c | c c}
\toprule[1pt]
Case & \shortstack{Distribution \\ Alignment} & \shortstack{Sampling \\ Pseudo Labels} & Accuracy & mAP \\ \hline
Only Pseudo Labeling &  &  & 55.20 & 71.98 \\
AdaMatch \cite{berthelot2021adamatch} & \checkmark &  & 68.01 & 82.13 \\
Semi-supervised Learning &  & \checkmark & 62.04 & 80.78 \\
Ours & \checkmark & \checkmark & \textbf{70.76} & \textbf{83.71} \\
\toprule[1pt]
\end{tabular}
\label{table:AB}
\end{table}

\textbf{Ablation Study.} We perform an ablation study to better understand the importance of each component in our method. In Table \ref{table:AB}, we analyze the effect of distribution alignment and sampling pseudo labels as the key components in our method. As we discussed in section \ref{sec:uda}, the classifier could predict the dominant class on target without these components. Although AdaMatch \cite{berthelot2021adamatch} uses distribution alignment to generate pseudo labels, they are usually unbalanced. We show that in datasets with a long-tail distribution, we also need to sample them to ensure more representative and balanced pseudo labels for training. We conduct our study in the UDA setting with OR1 used as source. Please see supplementary materials for more details.
% The supplementary material also includes the ablation experiments on different pseudo labeling strategies and the sensitivity analysis of hyperparameters such as $\lambda$ and batch size. 

\section{Conclusion}
In this paper, we studied the generalizability of surgical workflow analysis models as these models are known to suffer from a performance drop when deployed in a new environment. We proposed a new method for domain adaption, which relies on generating pseudo labels for the unlabeled videos from the target domain and training the model using the most confident ones. We showed that our method trains a more generalizable model and boosts the performance in both UDA and SSDA settings. Furthermore, our method is model-agnostics, as a result, it could be applied to other tasks using a suitable clip-based activity recognition and temporal model. For example, one of the areas is surgical activity recognition in endoscopic videos. We hope that our approach will inspire future work in medical machine learning to develop models that are more generalizable.

%
% ---- Bibliography ----
%
% BibTeX users should specify bibliography style 'splncs04'.
% References will then be sorted and formatted in the correct style.
%
\bibliographystyle{splncs04}
% \newpage
\bibliography{bibliography}

\newpage
\section*{Supplementary Material}

\subsection{Implementation Details}
In our experiments, we use the base TimeSformer model with hyperparameters described in \cite{bertasius2021space}. Our temporal model is a bidirectional GRU with $4$ hidden layers with the size of $256$ and a dropout of $0.1$. We train our models in Pytorch with an SGD optimizer and learning rate of $0.005$ for $30$ epochs. The learning rate for training the GRU model is set to $0.0005$. The hardware we used for training is $4$ Tesla A100 GPUs, each with $40$ GB of memory. Our clip-based activity recognition model takes around $2$ hours in the UDA setting, and our temporal model takes approximately $1$ hour to train. In SSDA settings, the model is trained on the source and the labeled part of the target, and it is tested on the whole target set. This brings the training time of the clip-based activity recognition model to around $4$ hours. 
\subsection{Pseudo Labeling Strategies}
In section \ref{sec:uda}, we described the pseudo labeling strategy with uniform confidence margin $\tau$, however, $\tau$ does not need to be fixed. \cite{berthelot2021adamatch} introduced relative confidence margin where $\tau$ is determined relative to the mean confidence of the top-1 predictions on the source. In our setting, it is defined as $\tau = \tau_0 \times \mathbb{E}_{p_{s} \sim \mathcal{P}_{s}} [\max(p_{s})]$.

In this work, we also propose an adaptive pseudo labeling strategies where the confidence margin $\tau$ is determined for each class separately. This is helpful in our settings where the dataset is imbalanced and the model is more confident in certain classes. Therefore, we define the confidence margin for class $i$ as
\begin{align*}
    \tau^{(i)} = \tau_0 \times \mathbb{E}_{\substack{p_{s} \sim \mathcal{P}_{s}\\ \hat{y}_s=i}} [\max(p_{s})]
\end{align*}
and the mask is defined as $mask = \max(\hat{p}_{t}) > \tau^{(\hat{y}_{t})}$.

We compare the performance of the pseudo labeling strategies in the UDA setting, where $\tau_0=0.9$. As shown in Table \ref{table:pseudo}, the uniform strategy has a better performance in the clip-based activity recognition model where the short clips are sampled uniformly from surgical actions. Adaptive pseudo labeling strategy lowers the $\tau$ for underrepresented classes, so more pseudo labels are generated for them. As a result, it performs better for the temporal model on our datasets with long-tailed distributions. We chose to use the uniform pseudo labeling strategy in the main text for simplicity.

\begin{table}[h]
\caption{Pseudo Labeling Strategies.}
\centering
\begin{tabular}{l | c c}
Case & Accuracy & mAP \\ \hline
Uniform & \textbf{70.76} & 83.71 \\
Relative \cite{berthelot2021adamatch} & 70.23 & 84.09  \\
Adaptive & 68.88 & \textbf{85.37} \\
\end{tabular}
\label{table:pseudo}
\end{table}

\newpage
\subsection{Sensitivity Analysis}
We measure the sensitivity of our method to hyperparameters. Table \ref{table:lambda} shows the effect of hyperparameter $\lambda$ on the performance of clip-based activity recognition and temporal models. As shown in the table, our model achieves a stable performance around $\lambda = 1$. However, very small $\lambda$ prevents the model from utilizing the unlabeled data, while very large $\lambda$ adds too much noise into the training as the model relies heavily on the generated pseudo labels. We set $\lambda = 1$ for the rest of the experiments. 

\begin{table}[h]
\caption{Sensitivity analysis on hyperparameter $\lambda$}
\centering
\begin{tabular}{l | c c}
$\lambda$ & Accuracy & mAP \\ \hline
0 & 62.13 & 76.39 \\
0.2 & 68.37 & 79.40 \\
0.4  & 70.32 & 81.23 \\
0.6  & 70.71 & 82.29 \\
0.8  & 69.85 & 82.56 \\
1.0 & \textbf{70.76} & 83.71 \\
1.2 & 70.66 & \textbf{84.05} \\
1.4  & 69.61 & 83.94 \\
1.6  & 69.36 & 83.87 \\
1.8  & 67.53 & 83.41 \\
5.0  & 64.45 & 83.29 \\
10  & 60.26 & 81.85 \\
\end{tabular}
\label{table:lambda}
\end{table}

We assume that the labeled and unlabeled batches have the same size throughout the main text. However, increasing the unlabeled batch size was shown to improve the performance \cite{berthelot2021adamatch}. We define $\beta$ as the ratio of unlabeled batch size to the labeled batch size. Table \ref{table:beta} shows the effect of $\beta$ on the performance of our method. For each $\beta$, we set the labeled batch size to the maximum that fits into the GPU memory. Therefore, for $\beta=1$ we have a batch of $16$ labeled video clips, and for $\beta=4$, we have a batch of $4$ labeled video clips. We see that increasing $\beta$ does not consistently help as we have to lower the batch size simultaneously. We set $\beta=1$ in other experiments as it has a lower computational cost.

\begin{table}[h]
\caption{Sensitivity analysis on hyperparameter $\beta$}
\centering
\begin{tabular}{l | c c}
$\beta$ & Accuracy & mAP \\ \hline
1 & \textbf{70.76} & 83.71 \\
2 & 68.73 & \textbf{84.31} \\
3  & 66.18 & 82.41 \\
4  & 69.41 & 81.35 \\
\end{tabular}
\label{table:beta}
\end{table}

\end{document}